\documentclass{article}

\usepackage[final]{tackling_climate_workshop_style}
\usepackage[utf8]{inputenc} % allow utf-8 input
\usepackage[T1]{fontenc}    % use 8-bit T1 fonts
\usepackage{hyperref}       % hyperlinks
\usepackage{url}            % simple URL typesetting
\usepackage{booktabs}       % professional-quality tables
\usepackage{amsfonts}       % blackboard math symbols
\usepackage{nicefrac}       % compact symbols for 1/2, etc.
\usepackage{microtype}      % microtypography
\usepackage{graphicx}
\usepackage{float}
\usepackage{amsmath}
\usepackage{tabularx}
\usepackage{multirow}
\usepackage{siunitx}
\usepackage{subcaption}

\title{Assessing Building Heat Resilience Using UAV and Street-View Imagery with Coupled Global Context Vision Transformer}

\author{%
  Steffen~Knoblauch \\
  Heidelberg University\\
  \texttt{steffen.knoblauch@uni-heidelberg.de} \\
  \And
  Ram~Kumar Muthusamy\\
  Heidelberg University\\
   \texttt{ramkumar.muthusamy@uni-heidelberg.de} \\
  \And
  Hao Li \\
  National University of Singapore \\
  \texttt{hao.li@nus.edu.sg} \\
  \And
   Iddy Chazua \\
   OpenMap Development Tanzania \\
  \texttt{iddy.chazua@omdtz.or.tz} \\
   \And
   Benedcto Adamu \\
   OpenMap Development Tanzania \\
  \texttt{benedcto.adamu@omdtz.or.tz} \\
   \And
  Innocent Maholi \\
  OpenMap Development Tanzania \\
  \texttt{innocent.maholi@omdtz.or.tz} \\
   \And
   Alexander Zipf \\
   Heidelberg University \\
  \texttt{zipf@uni-heidelberg.de} \\
}

\begin{document}

\maketitle

\begin{abstract}
Climate change is intensifying human heat exposure, particularly in densely built urban centers of the Global South. Low-cost construction materials and high thermal-mass surfaces further exacerbate this risk. Yet scalable methods for assessing such heat-relevant building attributes remain scarce. We propose a machine learning framework that fuses openly available unmanned aerial vehicle (UAV) and street-view (SV) imagery via a coupled global context vision transformer (CGCViT) to learn heat-relevant representations of urban structures. Thermal infrared (TIR) measurements from HotSat-1 are used to quantify the relationship between building attributes and heat-associated health risks. Our dual-modality cross-view learning approach outperforms the best single-modality models by up to 9.3\%, demonstrating that UAV and SV imagery provide valuable complementary perspectives on urban structures. The presence of vegetation surrounding buildings (versus no vegetation), brighter roofing (versus darker roofing), and roofing made of concrete, clay, or wood (versus metal or tarpaulin) are all significantly associated with lower HotSat-1 TIR values. Deployed across the city of Dar es Salaam, Tanzania, the proposed framework illustrates how household-level inequalities in heat exposure—often linked to socio-economic disadvantage and reflected in building materials—can be identified and addressed using machine learning. Our results point to the critical role of localized, data-driven risk assessment in shaping climate adaptation strategies that deliver equitable outcomes.
\end{abstract}

\section{Introduction} \label{chap:introduction}
Climate change is driving more frequent and intense heat waves, posing a growing public health threat \cite{Abunyewah.2025, Luthi.2023, Ebi.2021}. Urban heat islands, formed by human-built environments with extensive impervious surfaces, further exacerbate these risks \cite{Gao.2024}. Although heat exposure occurs both outdoors and indoors, the majority of daily heat burden is experienced indoors \cite{WhiteNewsome.2011, Amankwaa.2025}. Indoor heat resilience is strongly influenced by socio-economic factors, particularly in the Global South, where building insulation standards are often absent or inadequate. Consequently, socio-economically disadvantaged households disproportionately bear the health impacts of heat exposure despite contributing minimally to the underlying drivers of climate change \cite{Norton.2010, Yuan.2025}. Designing effective and equitable heat resilience strategies thus requires robust methods to monitor and characterize technical building attributes at scale.

Satellite imagery offers a promising avenue for large-scale extraction of technical building characteristics \cite{Zhu.2025}. While global building footprint datasets derived from satellite data provide extensive geometric information, they typically lack detailed attributes beyond basic shape and size. OpenStreetMap enables manual tagging of individual buildings, but its coverage and consistency are limited by the reliance on volunteer contributions. Recent studies highlight the value of SV imagery in capturing heat-relevant building features not visible from above, such as wall materials \cite{Tarkhan.2025, Xie.2025, Sun.2022b, Dai.2024, Dimitrov.2014}, colors \cite{Zhang.2021, Zhong.2021}, number of floors \cite{Li.2023}, and vegetation cover \cite{Seiferling.2017, Gong.2018}. Complementing this, UAV imagery—with its higher spatial resolution—outperforms satellite data in detailing roofing materials, often obscured in SV imagery, especially within dense urban canyons \cite{Knoblauch.2024, Kim.2021}. Open data platforms like OpenAerialMap and Panoramax facilitate the sharing and integration of such diverse UAV and SV datasets, advancing scalable mapping of technical building attributes. However, the combined use of these complementary data sources remains underexplored \cite{StarzynskaGrzes.2023, Zhang.2025}. To date, we are not aware of any study that has applied dual-modality learning approaches \cite{Hoffmann.2019} integrating aerial and SV imagery to extract building-specific features relevant to heat resilience. However, filling this research gap is essential for developing targeted and equitable climate mitigation strategies at the local scale.

\section{Materials and methods}\label{materials_and_methods}
\begin{figure}[H]
\centering
\includegraphics[width=1\textwidth]{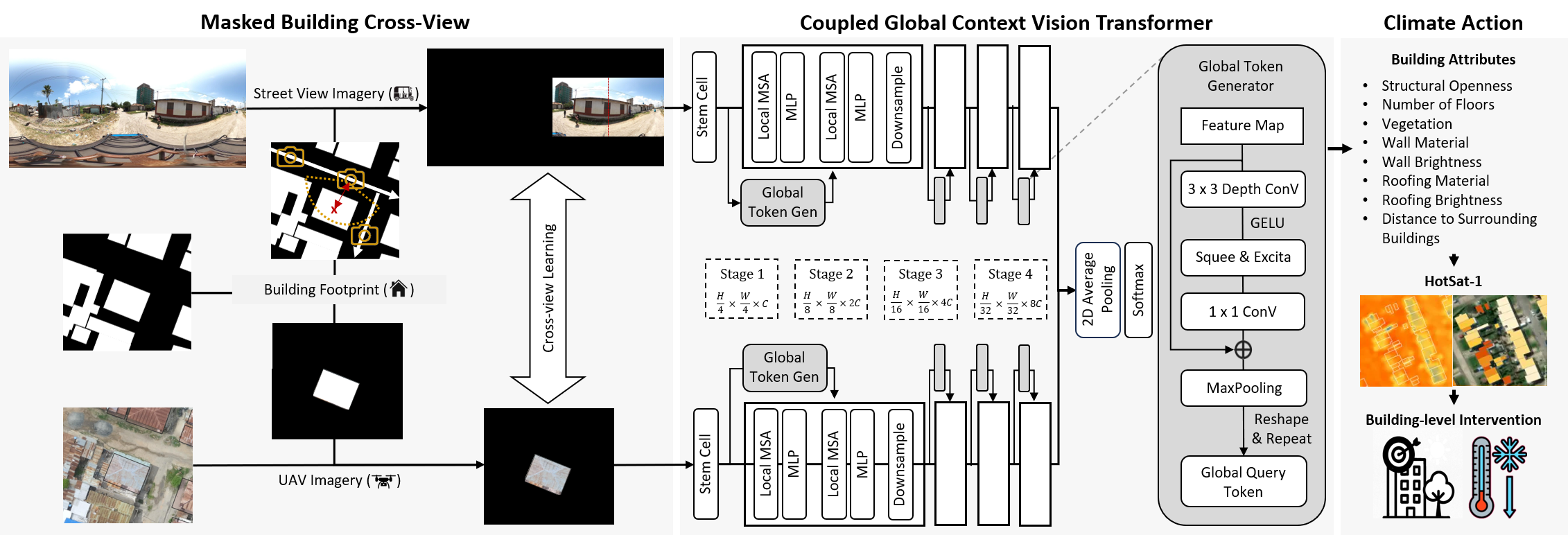}
\caption{Overview of the proposed CGCViT framework for extracting heat-relevant building attributes from paired UAV and SV imagery. Building footprints guide the extraction of cross-view image pairs, which are processed in parallel GCViT branches with global token integration. Predicted attributes include structural openness, number of floors, vegetation, wall/roofing material and brightness, and distance to surrounding buildings. These are statistically associated with  HotSat-1 TIR values to inform heat mitigation strategies at building-level.}
\label{fig:workflow}
\end{figure}

This study presents the development of a machine learning framework to extract heat-relevant building attributes from openly accessible geospatial datasets (cf. Fig. \ref{fig:workflow}). Inputs included (i) SV panoramic imagery, (ii) building footprints, and (iii) high-resolution UAV imagery to construct a cross-view representation of buildings. A CGCViT was trained to classify buildings by structural openness, number of floors, vegetation, wall material, and roofing material. These attributes, together with distance to surrounding buildings and roof and wall brightness, were statistically associated with HotSat-1 TIR values to identify priority targets for building-level heat mitigation by revealing features most strongly linked to lower thermal exposure. The framework was applied to the Msimbazi River Delta, Dar es Salaam, Tanzania—a densely populated floodplain (530,837 inhabitants) featuring compact urban development and diverse building typologies, providing an ideal context for investigating inequalities in urban heat exposure.

\subsection{Masked building cross-view}
SV imagery were collected via the Panoramax API in collaboration with OpenMap Development Tanzania (OMDTZ). A GoPro Max 360° camera mounted on a tricycle captured imagery at \num{2048} × \num{2048} pixels at sub-4-meter intervals along accessible roads (cf. Fig. \ref{fig:tricycle}). Data acquisition occurred in two phases: Oct 28–Nov 12, 2024, and Jan 6–Feb 8, 2025 \cite{OpenMapDevelopmentTanzania.2024}. Coverage was limited by inaccessible narrow footpaths in informal settlements and restricted institutional sites; these were excluded (cf. Fig. \ref{fig:challenges}). UAV imagery, sourced from OpenAerialMap and collected jointly with OMDTZ, was acquired using a DJI Mavic 2 Pro drone flying at 150 m altitude during a two-week campaign starting October 25, 2023 (cf. Fig. \ref{fig:study_area}). High-accuracy GPS ground control points ensured geospatial precision, yielding a 19.2 km² orthomosaic with 2.4 m horizontal and 1.4 m vertical accuracy and an average ground sampling distance of 9 cm. Building footprints from Geofabrik (downloaded April 8, 2025) were used to spatially align aerial and SV data across \num{63,844} buildings, of which \num{42,135} (\SI{65.98}{\percent}) were residential. Our analysis focused exclusively on residential structures, given their critical role in shaping household-level heat resilience. To ensure matching between aerial and SV imagery, the dataset was restricted to residential buildings with centroids within 30 m of a SV capture point and unobstructed frontal views, identified via a nearest-neighbor algorithm using building footprints. This filtering yielded \num{4,965} buildings with usable top- and front-view imagery (cf. Fig.\ref{fig:building-map}). To improve model focus, aerial and SV imagery were masked using building footprints. For SV imagery, driving direction metadata localized each building’s angular segment within the 360° image, enabling precise façade masking.

\subsection{Coupled global context vision transformer}
We trained the CGCViT on \num{2,000} masked cross-view building image pairs (\num{256}$\times$\num{256} px) with manual annotations for five classification tasks: structural openness, number of floors, surrounding greenery, wall material, and roofing material (cf. Fig. \ref{fig:example_annotations}, Tab. \ref{tab:label_stats_percent}). Wall and roof brightness were computed from mean RGB values of masked regions, and building density was estimated from mean distance to the four nearest OSM footprints (cf. Fig. \ref{fig:brightness_distributions}, Fig. \ref{fig:annotations}). The dataset was split into 70\%/15\%/15\% train/val/test, with data augmentation applied to minority classes. CGCViT processes UAV and SV imagery in parallel GCViT-Tiny \citep{Hatamizadeh.2022} streams, each with four stages of local and global self-attention \citep{Liu.2022,Liu.2021}, capturing fine-grained patterns and long-range dependencies (cf. Fig. \ref{fig:workflow}). A global query token is injected into local attention to aggregate context across regions via
\begin{equation}
\mathbf{G} = \text{Softmax}\Big( \frac{\mathbf{g}_q \mathbf{k}^\top}{\sqrt{s}} + \mathbf{p} \Big) \mathbf{v},
\end{equation}
where $\mathbf{g}_q$ is the global query, $\mathbf{k}$ and $\mathbf{v}$ are key and value matrices, $s$ is a scaling factor, and $\mathbf{p}$ is a learnable relative positional embedding. Features from both streams are concatenated, pooled, and classified using softmax cross-entropy
\begin{equation}
\mathcal{L}_{\text{softmax}} = - \sum_{c=1}^{C} y_c \log \frac{e^{z_c}}{\sum_{j=1}^{C} e^{z_j}}.
\end{equation}
For imbalanced tasks, focal loss was applied. Stratified 5-fold spatial cross-validation prevents geographic leakage. Model performance (support weighted-F1) was compared across UAV-only, street-view-only, and fused multi-modal cross-view learning settings. We assessed the statistical relationships between predicted building attributes and HotSat-1 TIR values (3.5\,m spatial resolution; acquired June–December 2023) using the Kruskal–Wallis test for categorical variables and Pearson’s correlation coefficient ($r$) for numerical variables.

\section{Results and discussion}
\begin{figure}[H]
\centering
\includegraphics[width=1\textwidth]{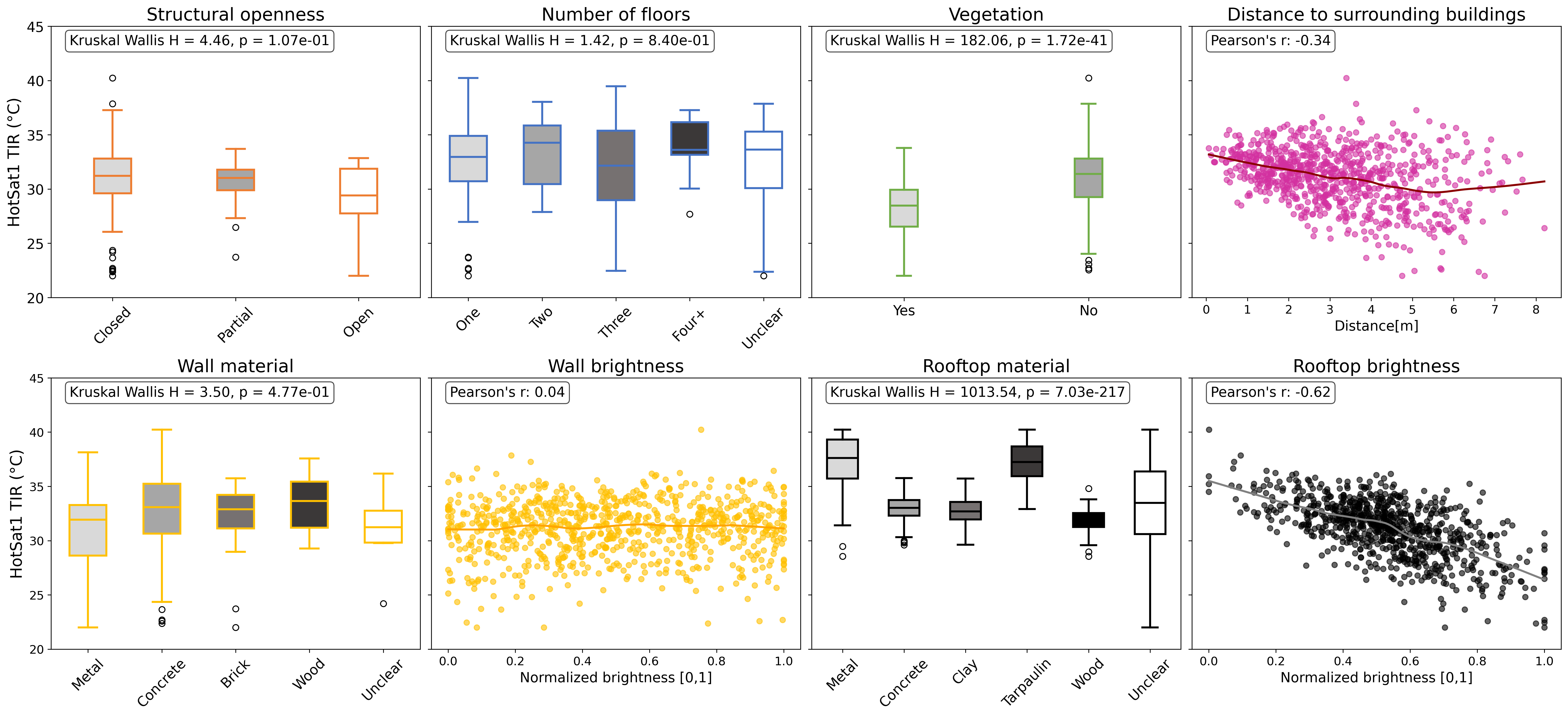}
\caption{Boxplots and scatter plots illustrate the relationships between HotSat-1 TIR values and building attributes identified in this study. They highlight a significant drop in mean HotsSat-1 TIR values for buildings surrounded by vegetation, as well as the important role of roofing materials and brightness.}
\label{fig:results}
\end{figure}

Cross-view learning using CGCViT consistently improved classification performance over the best single-modality models (UAV-only or SV-only), with gains in support weighted-F1 score ranging from 0\% to 9.3\% depending on the attribute. The most notable improvement occurred for vegetation classification (+9.3\%), followed by structural openness (+7.7\%) and number of floors (+7.1\%). Roofing material classification showed a modest gain (+3.7\%), while wall material classification exhibited no improvement over the best single-modality model (cf. Table \ref{tab:classification_f1_scores}). Vegetation classification benefited substantially from the multi-modal approach, as UAV imagery captures vegetation in backyards or on rooftops that are often out of reach for SV; conversely, SV imagery provides valuable views of vegetation located underneath shelters or on window ledges, complementing the UAV perspective. Number of floors was best predicted from SV imagery alone; however, incorporating UAV data via cross-view fusion yielded additional gains, likely because larger building footprints visible in UAV imagery tend to be associated with greater vertical building extent. Roofing material classification benefited markedly from UAV imagery compared to SV alone, but the combined multi-modal model achieved the highest accuracy. The improvement over the UAV-only model presumably reflects the predominance of low-rise buildings and wide streets in our study area, which make roofing details more discernible in SV imagery and thus provide complementary information to the UAV perspective—a condition that may not generalize to more complex urban environments. Wall material classification remains challenging, with lower accuracy overall, probably due to the near-complete invisibility of walls in UAV imagery and high rates of building obstruction in SV imagery (e.g., by vehicles, fences, or vegetation), both of which constrain the discriminative information available to single-modality models. For structural openness, modality-specific differences were minimal; nevertheless, the multi-modal model achieved modest F1-score improvements despite the inherent difficulty of the task, compounded by ambiguous class boundaries identified during annotation. Beyond classification accuracy, our analysis revealed significant associations between certain predicted building attributes and HotSat-1 TIR values (Fig. \ref{fig:results}). In particular, buildings with surrounding vegetation, roofing materials such as concrete, clay, or wood, and higher roof brightness exhibited lower mean TIR values, indicating a strong building-level cooling effect.

\section{Conclusion}
UAV and SV imagery, while not fully orthogonal, provide complementary perspectives that—when integrated via cross-view learning—improve the identification of heat-relevant building attributes. We find that greater vegetation cover, roofing materials such as concrete, clay, or wood, and higher roof brightness are consistently associated with lower HotSat-1 TIR values. This suggests that low-cost, household-level interventions—such as applying reflective roof coatings, providing seeds and water for small urban gardens, or supporting upgrades of heat-absorbing materials—can effectively reduce heat exposure. These scalable, machine-learning-derived insights enable targeted climate action while promoting equitable outcomes for households most vulnerable to climate-related impacts.

\bibliographystyle{unsrt}
\newpage
\bibliography{crossview_building_features}

\newpage

\appendix
\section*{Appendix}
%\subsection{Data Acquisition}

\begin{figure}[!ht]
\centering
\includegraphics[width=1\textwidth]{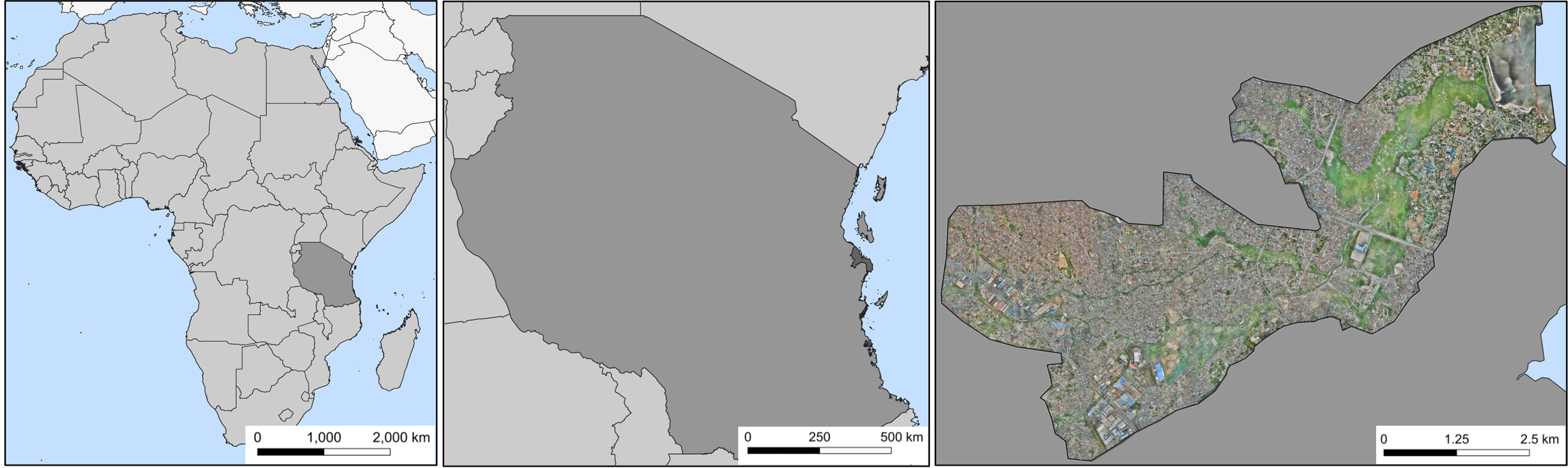}
\caption{Study area—the Msimbazi River Delta in Dar es Salaam, Tanzania—covered by UAV imagery used throughout the analysis.}
\label{fig:study_area}
\end{figure}

\begin{figure}[!ht]
\centering
\includegraphics[width=0.45\textwidth]{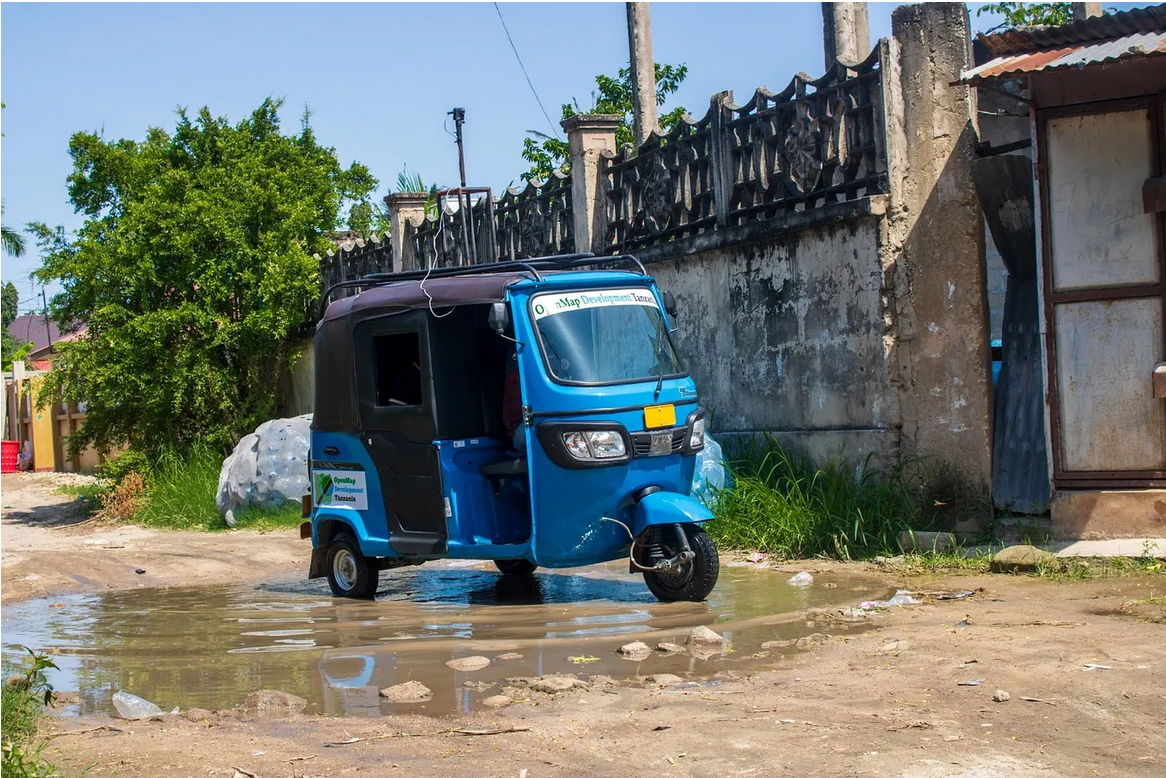}
\caption{Tricycle equipped with a GoPro Max used for collecting 360° SV imagery.}
\label{fig:tricycle}
\end{figure}

\begin{figure}[!ht]
    \centering
    \begin{subfigure}[t]{0.45\linewidth}
        \centering
        \includegraphics[width=\linewidth]{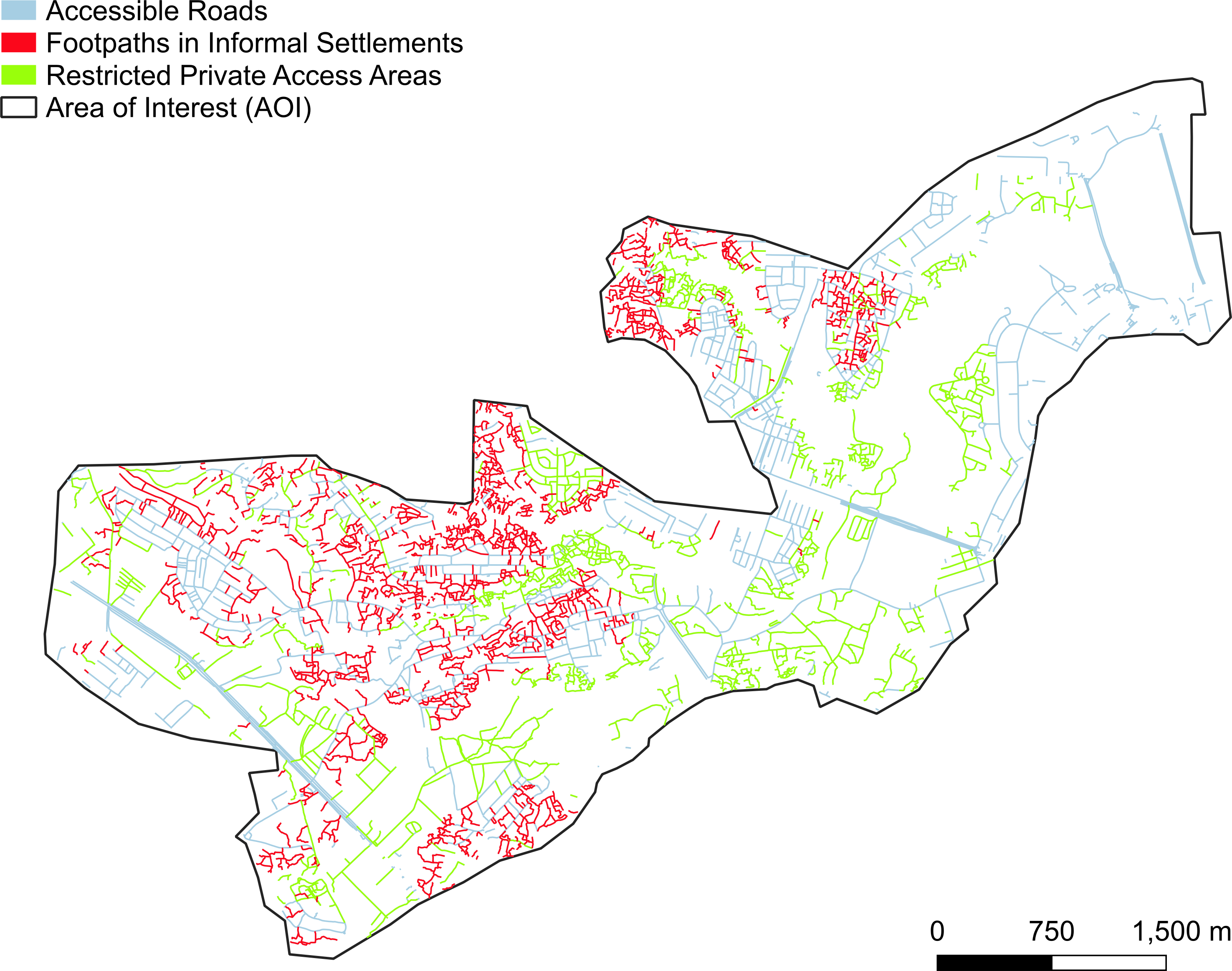}
        \caption{Map of the study area showing road accessibility for tricycle-based image collection. Navigable roads are marked in blue. Footpaths within informal settlements, which were inaccessible to the tricycle, are shown in red. Areas with restricted private access, such as institutional compounds, are indicated in green. This visualization highlights the spatial constraints affecting SV imagery coverage.}
        \label{fig:challenges}
    \end{subfigure}
    \hfill
    \begin{subfigure}[t]{0.45\linewidth}
        \centering
        \includegraphics[width=\linewidth]{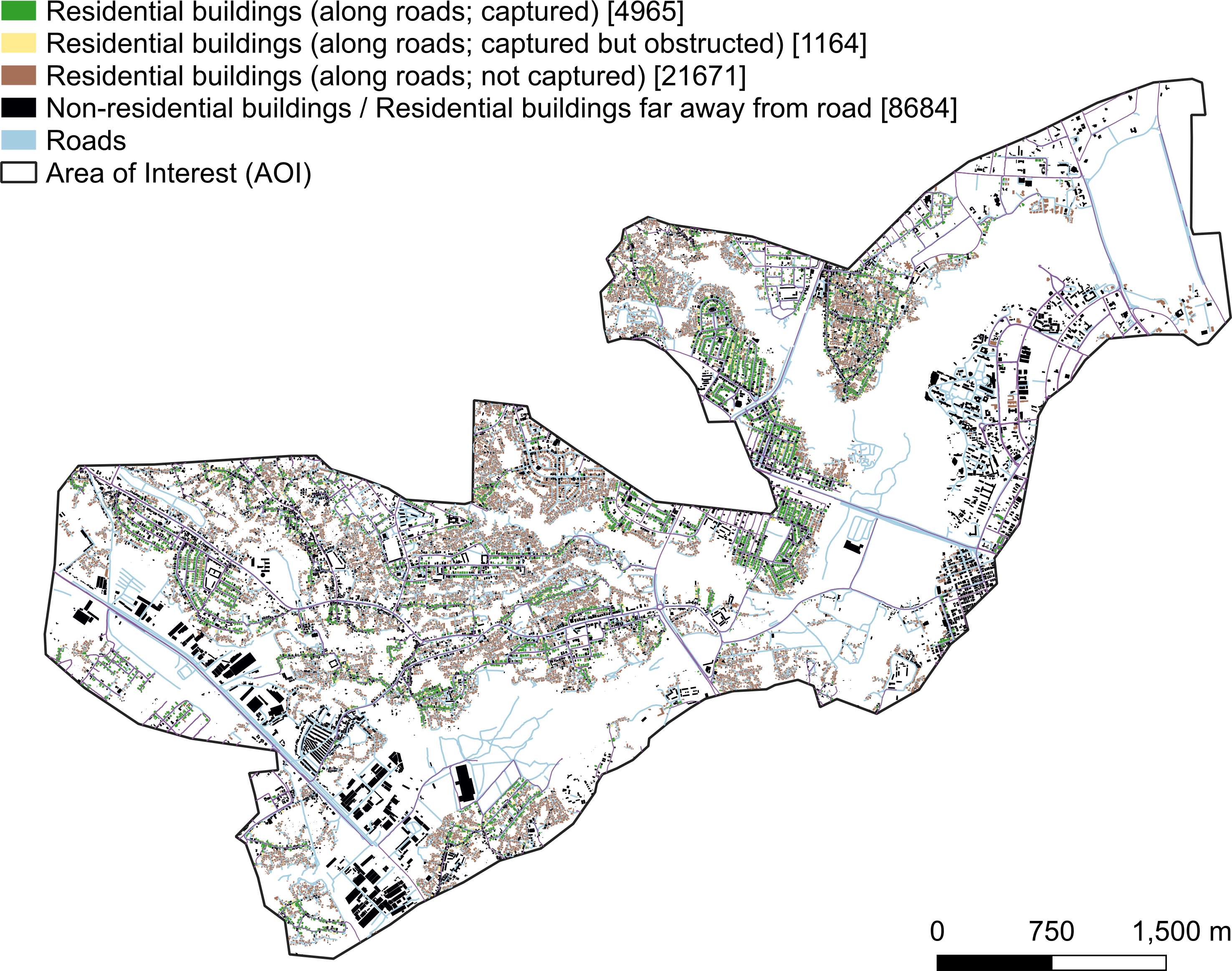}
        \caption{Map of the study area showing the road network (light blue), SV imagery capture points with 5-meter distance (purple), and building footprints. Black buildings are non-residential houses. Brown buildings represent residential structures whose footprint centroid lies more than 30\,m from the nearest image capture point. Yellow buildings are within 30\,m but are visually obstructed from the street (e.g., located behind other buildings). Green buildings meet all inclusion criteria and were used in the final analysis.}
        \label{fig:building-map}
    \end{subfigure}
    \caption{Data acquisition process and spatial coverage of SV imagery. (a) Tricycle-based setup used for image capture. (b) Spatial distribution and visibility classification of buildings in the study area.}
    \label{fig:figures_side_by_side}
\end{figure}

%\subsection{Data Annotation}
\begin{figure}[!ht]
    \centering
    \includegraphics[width=1\linewidth]{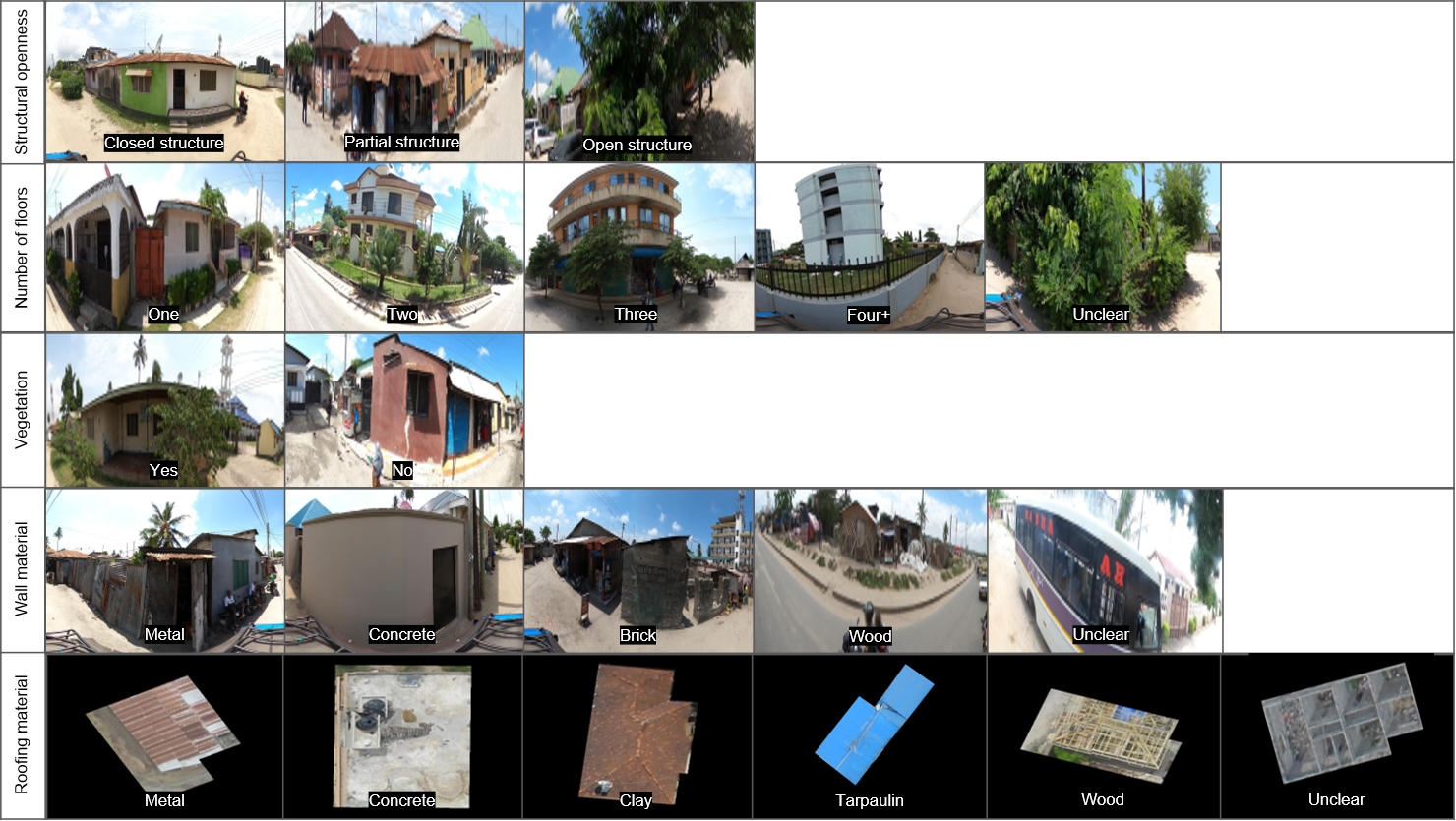}
    \caption{Exemplary manually annotated building images showing the range of class labels across different classification tasks.}
    \label{fig:example_annotations}
\end{figure}

\begin{table}[!ht]
\centering
\caption{Counts and percentage distribution of annotation classes.}
\label{tab:label_stats_percent}
%\footnotesize
\begin{tabular}{llrrr}
\toprule
\textbf{Classification task} & \textbf{Annotation Class} & \textbf{Count} & \textbf{Percentage (\%)} \\
\midrule
\multirow{4}{*}{Structural openness} 
 & Closed Structure & 1819 & 90.95 \\
 & Partial          & 44   & 2.20 \\
 & Unclear          & 137   & 6.85 \\

\midrule
\multirow{6}{*}{Number of floors} 
 & One       & 1902 & 95.10 \\
 & Two       & 34   & 1.70 \\
 & Three     & 11   & 0.55 \\
 & Four+      & 12   & 0.60 \\
 & Unclear   & 41   & 2.05 \\
\midrule

\multirow{2}{*}{Vegetation} 
 & Yes & 1946 & 97.30 \\
 & No  & 54   & 2.70 \\
\midrule
 
\multirow{5}{*}{Wall material} 
 & Metal     & 95   & 4.75 \\
 & Concrete  & 1820 & 91.00 \\
 & Brick     & 12   & 0.60 \\
 & Wood      & 5    & 0.25 \\
 & Unclear   & 68   & 3.40 \\
\midrule

\multirow{5}{*}{Roofing material} 
 & Metal     & 1744 & 87.20 \\
 & Concrete  & 22   & 1.10 \\
 & Clay      & 84   & 4.20 \\
 & Tarpaulin & 20   & 1.00 \\
 & Wood      & 63   & 3.15 \\
 & Unclear   & 67   & 3.35 \\

\bottomrule
\end{tabular}
\end{table}

\begin{figure}[!ht]
    \centering
    \begin{subfigure}[b]{0.49\textwidth}
        \centering
        \includegraphics[width=\textwidth]{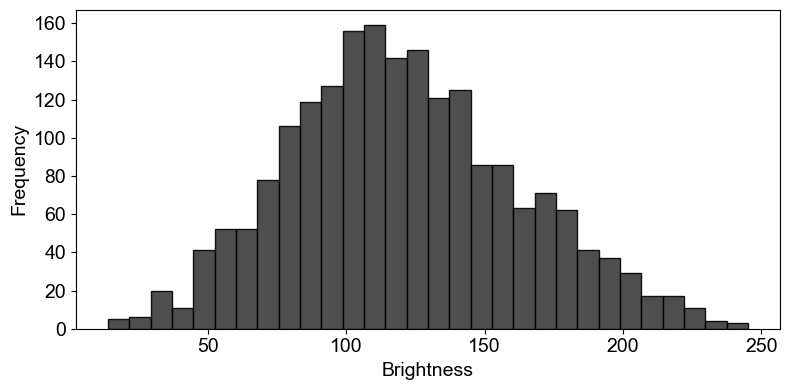}
        \caption{Wall}
        \label{fig:wall_albedo_dist}
    \end{subfigure}
    \hfill
    \begin{subfigure}[b]{0.49\textwidth}
        \centering
        \includegraphics[width=\textwidth]{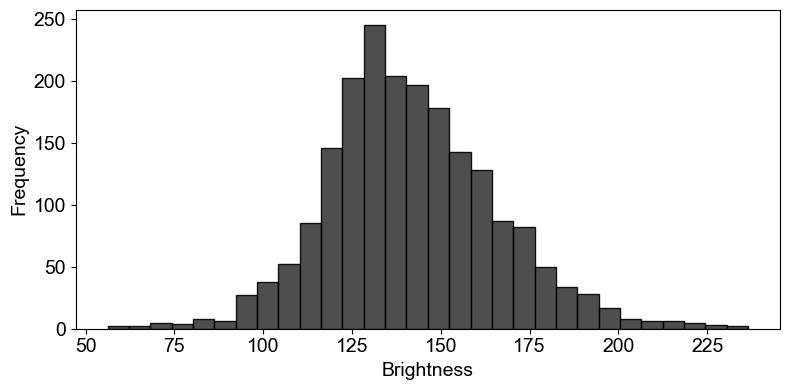}
        \caption{Roofing}
        \label{fig:sat_albedo_dist}
    \end{subfigure}
    \caption{Distribution of brightness values derived from RGB imagery, used as a proxy for surface reflectance and potential heat absorption characteristics.}
    \label{fig:brightness_distributions}
\end{figure}

\begin{figure}[!ht]
\centering
\includegraphics[width=1\textwidth]{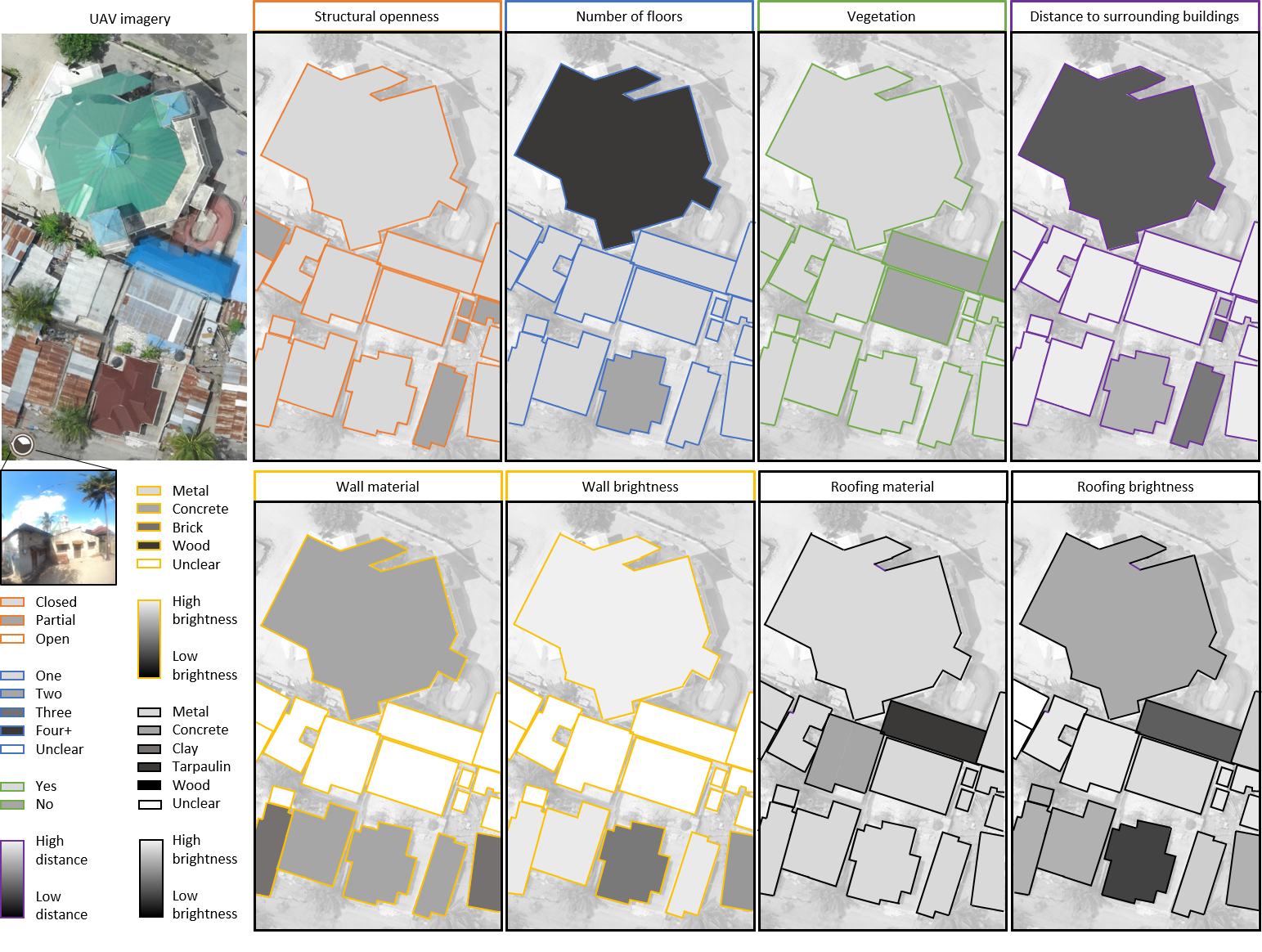}
\caption{Example of building-level annotations, based on UAV orthophotos and street-view imagery, corresponding to multiple heat-relevant building attributes. Building footprints are outlined and coloured by category: structural openness (orange), number of floors (blue), vegetation (green), distance to surrounding buildings (purple), wall material and brightness (yellow), and roofing material and brightness (black). Each of the eight panels on the right shows a separate attribute label overlaid on the same UAV imagery, enabling multi-label classification of individual buildings for cross-view learning. The imagery was captured near coordinates $-6.7985, 39.2686$.}
\label{fig:annotations}
\end{figure}

\clearpage
%\section{Results}
%\subsection{Cross-view learning}
\begin{table}[!ht]
\centering
\caption{Weighted F1-scores (weighted by class support) for classification tasks. Multi-modal results are compared against single-modality models using only UAV or SV data, showing the added value of cross-view learning for building attribute classification. Percentages indicate performance difference relative to multi-modal.}
\label{tab:classification_f1_scores}
\begin{tabularx}{\textwidth}{l *{3}{>{\centering\arraybackslash}X}}
\toprule
\textbf{Classification task} & \textbf{Multi-modal F1 (weighted)} & \textbf{SV F1 (weighted)} & \textbf{UAV F1 (weighted)} \\
\midrule
Vegetation                  & 0.94                              & 0.86 (-9\%)                        & 0.80 (-15\%)                       \\
Number of floors            & 0.91                              & 0.85 (-7\%)                        & 0.45 (-51\%)                       \\
Roofing material            & 0.85                              & 0.70 (-18\%)                       & 0.82 (-4\%)                       \\     
Wall material               & 0.68                              & 0.68 (+/-0\%)                      & 0.06 (-91\%)                       \\
Structural openness         & 0.66                              & 0.57 (-14\%)                       & 0.65 (-2\%)                       \\

%Vegetation                  & 0.94  & 0.86  & 0.80  \\
%Number of floors            & 0.91  & 0.85  & 0.45  \\
%Roofing material            & 0.85  & 0.70  & 0.82  \\     
%Wall material               & 0.68  & 0.68  & 0.06  \\
%Structural openness         & 0.66  & 0.57  & 0.65  \\

\bottomrule
\end{tabularx}
\end{table}

\end{document}